\documentclass[conference]{IEEEtran}
\IEEEoverridecommandlockouts
\usepackage{cite}
\usepackage{amsmath,amssymb,amsfonts}
\usepackage{algorithmic}
\usepackage{graphicx}
\usepackage{textcomp}
\usepackage{xcolor}
\usepackage{multirow}
\usepackage{amsmath}
\usepackage{amsfonts,color}
\usepackage{mathrsfs}
\usepackage{latexsym,amssymb}
\usepackage{graphicx}
\usepackage{cite}
\usepackage{subfigure,balance}
\usepackage{fancyhdr}  
\usepackage{graphicx}
\usepackage{algorithmic}
\usepackage{graphicx}
\usepackage{textcomp}
\usepackage{hyperref}
\usepackage{stfloats}
\usepackage{float}
\usepackage{amsmath}
\usepackage{amsfonts,color}
\usepackage{mathrsfs}
\usepackage{latexsym,amssymb}
\usepackage{graphicx}
\usepackage{cite}
\usepackage{subfigure,balance}
\usepackage{fancyhdr}
\usepackage{makecell}
\usepackage{booktabs}
\usepackage{algorithmic}
\usepackage{booktabs}
\usepackage{ragged2e}
\usepackage{tabularx}
\usepackage{bbding}
\usepackage{pifont}
\usepackage{wasysym}
\usepackage{amssymb}
\usepackage{float}
\usepackage{enumitem,amssymb}
\def\BibTeX{{\rm B\kern-.05em{\sc i\kern-.025em b}\kern-.08em
    T\kern-.1667em\lower.7ex\hbox{E}\kern-.125emX}}
\begin{document}

\title{A Lite Fireworks Algorithm with Fractal Dimension Constraint for Feature Selection \\

\thanks{This work was supported in part by the Key discipline promotion plan of computer science and technology (GKY-2020CQXK-2). }
\thanks{\textbf{Min Zeng and Haimiao Mo contributed equally to this work.}}
}

\author{\IEEEauthorblockN{1\textsuperscript{st} Min Zeng}
\IEEEauthorblockA{\textit{School of Computer Science,} \\
\textit{Guangdong University of Science and Technology,}\\
Dongguan, Guangdong, China. \\
Email: zm6102@163.com}
\and
\IEEEauthorblockN{2\textsuperscript{nd} Haimiao Mo*}
\IEEEauthorblockA{\textit{School of Management,} \\
\textit{Hefei University of Technology,}\\  Hefei, Anhui, China. \\
Email: mhm\_hfut@163.com}
\and
\IEEEauthorblockN{3\textsuperscript{rd} Zhiming Liang}
\IEEEauthorblockA{\textit{Department of Pharmacy,} \\
\textit{Fifth Affiliated Hospital of Guangzhou Medical University,}\\
Guangzhou, Guangdong, China.
}
\and
\IEEEauthorblockN{4\textsuperscript{th} Hua Wang}
\IEEEauthorblockA{\textit{School of Computer Science,} \\
\textit{Guangdong University of Science and Technolog,}\\
Dongguan, Guangdong, China.
}
}

\maketitle

\begin{abstract}
As the use of robotics becomes more widespread, the huge amount of vision data leads to a dramatic increase in data dimensionality. Although deep learning methods can effectively process these high-dimensional vision data. Due to the limitation of computational resources, some special scenarios still rely on traditional machine learning methods. However, these high-dimensional visual data lead to great challenges for traditional machine learning methods. Therefore, we propose a Lite Fireworks Algorithm with Fractal Dimension constraint for feature selection (LFWA+FD) and use it to solve the feature selection problem driven by robot vision. The "LFWA+FD" focuses on searching the ideal feature subset by simplifying the fireworks algorithm and constraining the dimensionality of selected features by fractal dimensionality, which in turn reduces the approximate features and reduces the noise in the original data to improve the accuracy of the model. The comparative experimental results of two publicly available datasets from UCI show that the proposed method can effectively select a subset of features useful for model inference and remove a large amount of noise noise present in the original data to improve the performance.

\end{abstract}


\section{Introduction}
As the demand for robotics continues to grow in various industries, the importance of developing robots that can learn and adapt to their environment is increasing \cite{javaid2021substantial}. However, a key challenge in creating such robots is processing and interpreting high-dimensional features of visual multimodal data \cite{alatise2020review}. Visual multimodal data is captured through multiple sensors, including cameras, microphones, and touch sensors, and can include features such as color, texture, shape, sound, and pressure. Deep learning methods have shown effectiveness in processing high-dimensional visual data \cite{esteva2021deep}, but due to limited computational resources, they still rely on traditional machine learning methods in some scenarios. Nonetheless, traditional machine learning techniques encounter significant challenges in processing high-dimensional visual multimodal data due to the increased data dimensionality.

Feature selection \cite{solorio2020review} is a critical step in machine learning that aims to improve model accuracy, reduce overfitting, and decrease computation time by selecting a subset of relevant features from a larger set of features. It is particularly important when working with high-dimensional datasets, where using all available features may lead to overfitting and reduce the model's generalization ability. Various techniques for feature selection exist, including filter methods \cite{bommert2020benchmark}, wrapper methods \cite{al2022wrapper}, and embedded methods \cite{nguyen2020survey}.

%
%
%

Each method has its own advantages and disadvantages, and the choice of a feature selection method depends on the characteristics of the dataset and the goal of the study. The wrapper method can provide better accuracy, but it is computationally expensive and can overfit. The filter method is computationally inexpensive, but it may select irrelevant features. The embedded method is a hybrid method that can provide a balance between wrapper and filter methods. In recent years, because the search ability of evolutionary algorithm \cite{zhang2014hybrid}, \cite{lapa2016application}, such as fireworks algorithm \cite{li2019comprehensive}, can effectively avoid falling into the local optimal solution and optimize non-differentiable problems, the embedded feature selection method based on swarm intelligence algorithm has attracted much attention.

To address the needs of scenarios where robots acquire large amounts of data in scenarios with limited computational resources that make traditional machine learning methods inapplicable, we propose a Lite Fireworks Algorithm with Fractal Dimension constraint for feature selection (LFWA+FD).

The contributions of this paper are as follows.

\begin{itemize}
	\item 
	
	We propose an embedded feature selection framework that approximates features by fractal dimensionality and searches for feature subsets by the Lite Fireworks Algorithm to reduce data noise and data dimensionality, thereby reducing computational resource constraints and improving model accuracy. 
	
	\item 
	We propose a Lite Fireworks Algorithm which reduces the parameter tuning cost of the original fireworks algorithm by constructing new explosion operators, including explosion intensity and explosion radius.

\end{itemize}

\section{Related Work}

\subsection{Feature Selection based on Evolutionary Algorithm}

Unsupervised approaches are commonly used in engineering to reduce feature noise, as they offer the advantages of low overfitting risk and high performance. The three primary techniques employed in unsupervised feature selection (UFS) are wrapper, filter, and embedded methods \cite{solorio2020review}. Wrapper and embedded methods typically rely on the global exploration and local mining capabilities of evolutionary algorithms to identify optimal feature subsets.

Wrapper methods have several drawbacks, including high computational cost, susceptibility to overfitting, and the requirement of using specific clustering algorithms. According to the feature search strategy, the UFS technique based on the wrapper method can be classified into three categories: sequential, bio-inspired, and iterative \cite{solorio2020review}. In the sequential method, features are added or removed sequentially, making it a quick and straightforward technique. The bio-inspired approach, on the other hand, introduces randomness to avoid local optimum results and is often based on the search ability of evolutionary algorithms such as Particle Swarm Optimization (PSO) \cite{song2021fast}, Bat Algorithm (BA) \cite{yu2020chaos}, Artificial Bee Colony (ABC), and Ant Colony Optimization (ACO)  \cite{ghosh2020wrapper}. In iterative approaches, the unsupervised feature selection problem is transformed into an estimation problem, eliminating the need for combinatorial searches.

Filter methods determine the most pertinent features by analyzing the data itself and do not require clustering algorithms to direct the search. These methods evaluate features based on data properties such as distance, informativeness, independence, and saliency tests \cite{bommert2020benchmark}. Filter methods' speed and scalability are significant advantages.

Embedded techniques combine the advantages of filtering and wrapper methods \cite{nguyen2020survey}. These methods use measures to rank or select features in the filtering phase based on the data's inherent properties. In the wrapper stage, various feature subsets are compared using a specific clustering algorithm to identify the best selection. Embedded methods consist of two stages, where the filter method initially removes irrelevant or noisy features, and only important features are retained. Then, the remaining features and sample data are passed to the wrapper selection method as input parameters to optimize the selection. This approach reduces dimensionality and improves classification accuracy.

\subsection{Improvement Works of Fireworks Algorithm}

The Fireworks Algorithm (FWA) \cite{li2019comprehensive}, which Tan et al. first announced in 2010, was inspired by the way that fireworks naturally erupt in the night sky to create sparkles and illuminate the surroundings. Because of its quick convergence, high precision, and ability to balance local mining and global exploration through exploding sparks, the fireworks algorithm is frequently employed in function optimization \cite{cheng2019improved}, pattern recognition \cite{zhang2020recognition}, optimum scheduling \cite{he2019discrete}, and other disciplines.

According to earlier investigations, the original fireworks algorithm has certain design flaws, according to earlier investigations \cite{li2019comprehensive}. First, every dimension causes the same displacement when the original fireworks algorithm explodes. It might cause the population to be too homogeneous, which may restrict the algorithm's capacity for exploration. Second, the original fireworks choose the offspring fireworks using a distance-based selection technique, which results in ineffective offspring fireworks selection and higher computing costs. Thirdly, the original firework algorithm contains a variety of subpopulations, including the fireworks subpopulation, the explosion spark subpopulation, the explosion spark subpopulation, and the Gaussian spark subpopulation. The fireworks algorithm's local mining and global exploration capacities are constrained by the lack of information-sharing and coordination mechanisms among these subpopulations.

Based on the drawbacks of the original fireworks algorithm, the improvement efforts for this algorithm are mainly divided into two categories \cite{zhigang2022review}. The first category of improvement work focuses on enhancing mutation operators, mapping rules, selection techniques, and explosion operators (such as explosion intensity, explosion radius, and displacement operations). Enhancing the fireworks algorithm by fusing the benefits of different heuristic algorithms is another category of improvement effort \cite{li2019comprehensive}.

There are three basic improvement tactics used in the first type of improvement work. First, the fireworks algorithm's local mining and global exploration capabilities (Increasing  population diversity) are increased by utilizing the effective information of fireworks (mapping rules, or advantages and disadvantages of an explosion spark difference vector). The enhancement technique effectively boosts the fireworks algorithm's potential for optimization by enhancing population diversity \cite{li2015orienting}, random mapping rules \cite{cheng2015analytics}, \cite{ye2018mapping}, and explosion radius \cite{li2014adaptive}, \cite{zheng2014dynamic}, \cite{zheng2015cooperative}, \cite{pekdemir2016enhancing}. Second, this kind of improvement approach is mostly enhanced for the fireworks algorithm's offspring selection strategy \cite{pei2012empirical}, \cite{jun2019fireworks}, \cite{chen2021explosion}, such as "random-elite" strategy and tournament strategy, because to the low efficiency and high cost of selecting offspring. Third, through the transmission of information amongst sub-populations, the enhanced firework algorithm creates a powerful collaborative optimization mechanism between sub-populations \cite{zheng2015cooperative}, \cite{lv2022novel}, \cite{liu2021neighborhood}, \cite{li2016enhancing}, \cite{li2006tan}. It may successfully increase the algorithm's capacity for optimization while also enhancing the effectiveness of information usage among the subpopulations of the original fireworks algorithm and choosing top-notch subpopulations to lead the population toward the best solution.

The second type of improvement work introduces the advantages of other evolutionary algorithms such as  Biogeography-based Optimization Algorithm (BOA) \cite{zhang2014hybrid}, and Genetic Algorithm (GA) \cite{lapa2016application} to enhance the performance of the hybrid algorithm. By incorporating the advantages of these evolutionary algorithms, the local mining and global exploration capacities of the original fireworks method are considerably improved.

However, the majority of current research focuses on fixing fireworks algorithm flaws rather than lowering the method's extensive amount of parameters \cite{li2019comprehensive}. The original fireworks method needs to utilize a constant to regulate the number of explosion sparks while computing the explosion intensity.

\section{Feature Selection's Framework}

\subsection{The Lite Fireworks Algorithm}

The majority of current research focuses on fixing fireworks algorithm flaws rather than lowering the method's extensive amount of parameters \cite{li2019comprehensive}. To calculate the explosion radius, the maximum explosion radius threshold needs first to be defined in the original fireworks algorithm. The original fireworks algorithm has an excessive number of parameters, which increases the cost of algorithm tuning. Additionally, individuals whose offspring fireworks are not chosen for the next generation will be deleted, which significantly wastes the history information of these individuals. We construct an adaptive explosion radius that balances the fireworks algorithm's local mining and global exploration capabilities by reducing the complexity of the algorithm's parameters and completely utilizing the previous data of the unselected individuals.

\subsubsection{Explosion Intensity}

The original fireworks method needs to specify the maximum number of explosion sparks beforehand, therefore having too many parameters may make it difficult to use.  The new way of generating explosion intensity, which reduces the number of parameters in the fireworks method, is illustrated by Equation (1). The number of explosion sparks that can occur is thus limited by the population size and fitness value of each individual firework.

\begin{equation}
	{S_i} = \left\lceil {{M^{\frac{{{f_{\max }} - f({x_i})}}{{{f_{\max }} - {f_{\min }} + \xi }}}}} \right\rceil
\end{equation}
where $S_i$ is the \textit{i}-th fireworks $x_i$'s explosive intensity, the population of fireworks is $M$. The ceil function $\left\lceil {} \right\rceil $ is used to restrict the number of $S_i$ in the integer range $[1, M]$. $\xi$ is a very small constant used to limit the denominator to zeros. When optimizing the minimum objective function, $f_{max}$ and $f_{min}$ represent the fitness values of the worst and best positions of the contemporary population respectively. $f(x_i)$ represents the fitness value of the \textit{i}-th firework.

\subsubsection{Explosion Radius}

The history data of the candidates who were not chosen will be greatly wasted because the original fireworks algorithm would reject them when choosing the next generation of fireworks. In order to balance the local mining and global exploring capabilities of the fireworks algorithm, a new explosion radius is constructed utilizing the historical data of each of these individual locations. Equations (2) and (3) provide the updated explosion radius calculations (3).

\begin{equation}
	{R_i} = \left\{ \begin{array}{l}
		pbes{t_i} - {x_i}, {S_i} < {S_{Avg}}\\
		x_{CF} - x{_i}, {S_i} \ge {S_{Avg}}
	\end{array} \right.
\end{equation}
\begin{equation}
	{S_{Avg}} = \frac{1}{M}\sum\limits_{i = 1}^M {{S_i}} 
\end{equation}
where the \textit{i}-th fireworks $x_{i}$'s historical best position is indicated by $pbest_{i}$, and $i=1,2,...,M$. $M$ is the size of the fireworks population. The best firework individual  $x_{CF}$  is selected from $pbest = \{ pbes{t_1},pbes{t_2},...,pesbt{}_M\}$, commonly known as Core Fireworks (CF). $S_{Avg}$ is the mean explosion intensity, which is also the mean number of all sparks explosion sparks.

\subsubsection{Displacement Operation}
The \textit{i}-th fireworks $x_i$ finish the displacement when they explode, and then $S_i$ are produced as explosion sparks. Equation (4) provides the displacement function.

\begin{equation}
	E{S_k} = {x_i} + \beta  \times {R_i}
\end{equation}
where $k=1,2, ..., S{_i}$. And $x_i$ is the position of \textit{i}-th fireworks, and $ES_k$ is the \textit{k}-th Explosion Sparks (ES) in the \textit{i}-th subpopulation generated by the \textit{i}-th fireworks $x_i$ after an explosion. $\beta$ has a uniform distribution and is generated at random between [0, 1], $R_i$ is the \textit{i}-th fireworks $x_i$'s explosion radius.

Equations (2) to (4) are used by LFWA to adopt $x_{CF}$ and $pbest$ to create a new explosion radius that can be adjusted adaptively based on the explosion intensity. If the average explosion intensity of a firework $S_{Avg}$ is less than the average explosion intensity of the \textit{i}-th firework $x_i$, the firework learns from $pbest$. In any other case, the fireworks pick up knowledge from the core fireworks $x_{CF}$. By including $x_{CF}$ and $pbest$ to build a new explosion radius, LFWA can change the step size flexibly.

\subsubsection{Mutation Factor}

In order to limit the population of fireworks from too quickly approaching the local optimal solution, the LFWA employs the mutation mechanism to maintain diversity in the fireworks population. Equation (5) illustrates the LFWA's mutation strategy, which involves randomly choosing fireworks' \textit{n}-dimensions for Gaussian mutation. After the fireworks undergo the Gaussian mutation, Gaussian Sparks (GS) are produced.

\begin{equation}
	G{S_{ij}} = {x_{ij}} \times (N(0,1) + 1)
\end{equation}
where $j =1,2,..., n$. $N(0, 1)$ is a Gaussian distribution function with a mean and standard deviation of 0 and 1, respectively. The total dimensions of $x_i$ are $d$.

\subsubsection{Mapping Rules}

To avoid out-of-bounds situations, it is necessary to process each dimension of the positions of the fireworks, explosion sparks, and Gaussian sparks using the mapping rules of Equation (6).

\begin{equation}
	{x_{ij}^{new}} = LB + \beta  \times (UB - LB)
\end{equation}
where the search range of the algorithm has upper and lower bounds, respectively, denoted by $UB=1$ and $LB=0$.

\subsubsection{Selection Strategy}

The LFWA selects the fireworks' offspring using the "Elite-Random" technique from a candidate set that comprises fireworks $x = \{ x{}_1,{x_2},...,x{}_M\}$, $pbest$, $x_{CF}$, explosion sparks $ES$, and Gaussian sparks $GS$. In other words, the core fireworks individual with the best fitness is chosen to symbolize the next generation of fireworks individuals, and the other fireworks individuals are chosen at random.

\subsection{LFWA-based Feature Selection and Inference}

\begin{figure}[htbp]
	\begin{center}
	\includegraphics[width = 8 cm]{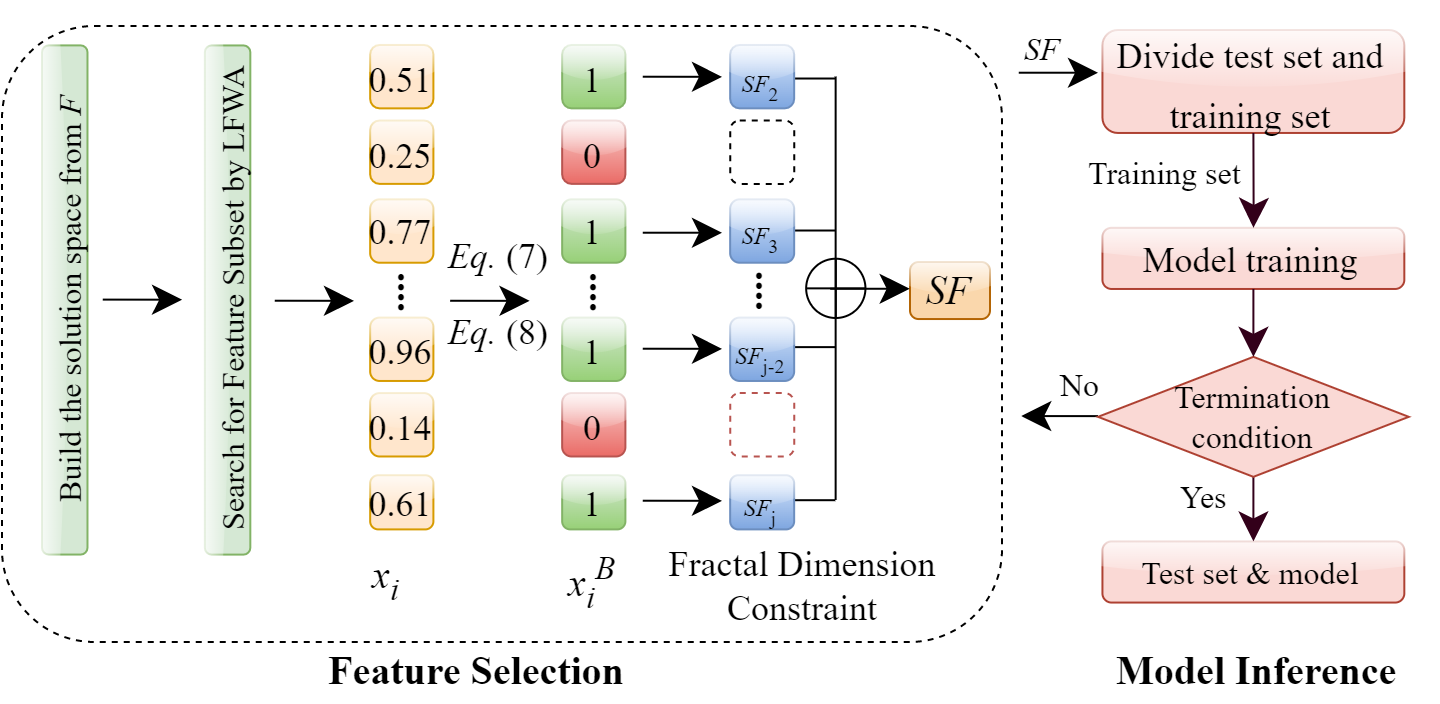}
	\end{center}
	\caption{The framework of LFWA-based feature selection and model inference}
	\label{Fig1}
\end{figure}

The process of selecting a feature subset can be viewed as an optimization 0-1 programming problem. The total dimensions of features are also $d$. Then, there are a total of $2^d$ different combinations of feature subsets, which constitute a solution space. The LFWA-based feature selection method is an embedded feature selection method. In order to find an ideal feature subset in the solution space, the method searches for the location where the feature subset is located, which in turn leads to the optimal performance of the model. The selected feature dimensions are constrained by Fractal Dimension \cite{nayak2019analysing}.

Figure 1 shows the framework of LFWA-based feature selection and model inference. First, a solution space composed of $2^d$ different subsets of features is constructed from the features $F$. Then, the feature selection method based on LFWA searches for the position of the feature subset. In other words, each firework individual $x_i$ in the iterative process of the LFWA algorithm represents a set of decision variables, that is, the position $x_{i}^{B}$ of a feature subset. $x_{i}^{B}$ is the result of $x_i$ being \textit{Binarized} by Equation (7). The selected subset of features is determined by position $x_{i}^{B}$. For example,  $x_{ij}^{B}$=1 means that the feature of the \textit{j}-th dimension in the feature $F$ is selected, otherwise it is not selected. The number of selected feature dimensions is limited by the Fractal Dimension. Next, the Selected Features (SF) are divided into a test set and a training set. The training set is used to train the model until the termination condition is reached, such as the preset maximum number of iterations, then the feature selection is terminated. The trained model and test set are used to make predictions. Otherwise, the LFWA-based feature selection method continues to search for a new feature subset.
\begin{equation}
	x_{ij}^B = \left\{ \begin{array}{l}
		0, {x_{ij}}\, < 0.5\\
		1, {x_{ij}} \ge 0.5
	\end{array} \right.
\end{equation}
where $x_{ij}$ is the value of the \textit{j}-th dimension of the \textit{i}-th firework individual $x_{i}$, and $j=1,2,...,d$.

Fractal means that the parts of a system are similar to the whole in some way. Fractal Dimension (FD)\cite{nayak2019analysing} is the main parameter describing the fractal and also is a statistic that indicates how well the fractal fills the corresponding embedding space, and its magnitude portrays the complexity of the system. Fractal dimension, as a characteristic statistic reflecting the distribution of a data set, has been widely used in the field of data mining. The calculation methods of fractal dimension include box calculation method \cite{so2017enhancement}, Higuchi method \cite{siamaknejad2019fractal}, wavelet transform method \cite{sharma2017new}, etc. The box calculation method is used to calculate the FD by Equation (8).

\begin{equation}
	FD = \mathop {\lim }\limits_{n \to \infty } \frac{{\ln \sum\limits_i {C_{r,i}^2} }}{{\ln r}},r \in [{r_1},{r_2}]
\end{equation}
where ${C_{r,i}}$ is the number of data points falling into the \textit{i}-th lattice,  ${r}$ is the radius of the lattice, and ${\left[ {{r_1},{r_2}} \right]}$  is the scale-free space of the dataset.

Finally, the LFWA-based feature selection method optimizes the objective function of Equation (9).
\begin{equation}
	F({x_{ij}^B}) =  - max(Acc),\sum\limits_{j = 1}^d {x_{ij}^B}  = \left\lceil {FD} \right\rceil
\end{equation}
The meaning of $\left\lceil {} \right\rceil$  in Equation (9) is the same as that of Equation (1). The accuracy (Acc) of the model is maximized when the selected feature subset is used as the input to the classification model, such as K-nearest Neighbors algorithm (KNN) \cite{pan2020new}.

\section{Experiments}

\subsection{Parameter settings}

The population size of fireworks is set to 5, the total number of sparks is set to 50, the number of Gaussian sparks is set to 5, the maximum number of evaluations is set to 200, and the experiment is repeated independently for 20 times.

\subsection{Feature Reduction}

\begin{table}[H]
	\centering
	\caption{Description of the UCI dataset used in the experiments}
	\scalebox{0.95}{\begin{tabular}{ccccc}
			\toprule
			\multicolumn{1}{l}{\multirow{2}[2]{*}{Datasets}} & \multicolumn{1}{l}{\multirow{2}[2]{*}{Dataset name}} & \multicolumn{1}{l}{\multirow{2}[2]{*}{Rows}} & \multicolumn{1}{l}{Feature  } & \multicolumn{1}{l}{\multirow{2}[2]{*}{Categories}} \\
			&       &       & \multicolumn{1}{l}{dimension} &  \\
			\midrule
			\multicolumn{1}{l}{Dataset1} & \multicolumn{1}{l}{Image Segmentation} & \multicolumn{1}{l}{2310} & \multicolumn{1}{l}{19} & \multicolumn{1}{l}{7} \\
			\multicolumn{1}{l}{Dataset2} & \multicolumn{1}{l}{Spectf} & \multicolumn{1}{l}{267} & \multicolumn{1}{l}{44} & \multicolumn{1}{l}{2} \\
			\midrule
			\multicolumn{5}{c}{UCI datasets are from \textit{http://archive.ics.uci.edu/ml/datasets.php}.} \\
		\end{tabular}%
	}
	\label{tab:table1}%
\end{table}%

\begin{table}[H]
	\centering
	\caption{Results of feature reduction by the Fractal Dimension method}
	\scalebox{1}{\begin{tabular}{llll}
			\toprule
			Datasets & Fractal dimension& $\left\lceil {FD} \right\rceil$ & Reduction rate (\%) \\
			\midrule
			DataSet1 & 2.3965 & 3 & 84.21 \\
			DataSet2 & 2.6869 & 3 & 93.18 \\
			\bottomrule
		\end{tabular}%
	}
	\label{tab:table2}%
\end{table}%

\begin{table}[H]
	\centering
	\caption{The results of the comparison methods}
	\scalebox{0.98}{	\begin{tabular}{lllll}
			\toprule
			Datasets & Methods & Feature subset & Acc (\%) & $\Delta$Acc (\%)\\
			\midrule
			\multirow{5}[2]{*}{Dataset1} & FSA+RS & \{1, 2, 13\} & 84.02  & +9.89\\
			& AFSA+RS &\{1, 2, 10\}  & 87.00 & +6.91 \\
			& BGSO+FD & \{1, 2, 15\} & 77.64 & +16.27\\
			& IDGSO+MFD & \{1, 2, 15\} & 77.64 &+16.27 \\
			& FWA+FD & \{2, 12, 16\} & 90.62 & +3.29 \\
			& \textbf{LFWA+FD} & \{2, 11, 13\} & \textbf{93.91}  & - -\\
			\midrule
			\multirow{5}[2]{*}{Dataset2} & FSA+RS & \{3, 20, 38, 43\} & 71.23  & +7.76\\
			& AFSA+RS & \{19, 26, 42, 43\} & 73.08  & +5.91\\
			& BGSO+FD & \{2, 26, 41\} & 73.08  & +5.91\\
			& IDGSO+MFD & \{26, 40, 43\}& 73.18 & +5.81 \\
			& FWA+FD & \{34, 36, 40\} & 76.23 & +2.76\\
			& \textbf{LFWA+FD} & \{9, 26, 38\} & \textbf{78.99} & - -\\
			\bottomrule
		\end{tabular}%
	}
	\label{tab:table3}%
\end{table}%

\begin{table*}[h]
	\centering
	\caption{Comparison results of original features and reduced features}
	\scalebox{0.96}{	\begin{tabular}{llllll}
			\toprule
			Datasets & Methods & Feature subset & Feature dimensions & Acc (\%)  & $\Delta$Acc (\%) \\
			\midrule
			\multirow{3}[2]{*}{Dataset1} & M1    & \{1, 2, 3, 4, ..., \textit{d}-1, \textit{d}\}   & \textit{d}  & 95.72 & -1.81 \\
			& M2  & \{2, 4, 6, 8, 11, 14, 16\}  & 7 & \textbf{97.05} & -3.14 \\
			& \textbf{LFWA+FD} & \{2, 11, 13\}  & 3 & 93.91 & - - \\
			\midrule
			\multirow{3}[2]{*}{Dataset2} & M1    & \{1, 2, 3, 4, ..., \textit{d}-1, \textit{d}\}   & \textit{d} & 74.14 & +4.85 \\
			& M2  & \{2, 5, 6, 7, 11, 13, 15, 16, 18, 21, 23, 26 28, 30, 31, 32, 34, 35, 36, 38, 40, 44\}  & 22 & \textbf{83.03} & -4.04 \\
			& \textbf{LFWA+FD }& \{9, 26, 38\}  & 3 & 78.99 & - -\\
			\bottomrule
		\end{tabular}%
	}
	\label{tab:table4}%
\end{table*}%

Table 1 shows the UCI datasets used in the comparison experiments. These datasets can be accessed by automated machine equipment such as robots in different application scenarios. 

In this subsection of the experiments, the features of the original data are dimensionally reduced only by the Fractal Dimension method. The experimental results of the features' Reduction Rate ($RR$) are shown in Table 2. $RR$ is calculated by ${\rm{RR}} = \frac{{d - \left\lceil {FD} \right\rceil }}{d} \times 100\%$. \textit{d} is the dimension of the original feature. The results in Table 2 show that the fractal dimension method can greatly reduce the feature dimension of the original data with the Reduction Rate $RR$=84.21\% and $RR$=93.18\%, respectively. It also describes the original data with fewer features.

\subsection{Comparative Experiments}

In this set of experiments, we use the fractal dimensions from Table 2 to constrain the dimensions of the Selected Features $SF$. In order to verify the effectiveness of the feature selection method based on fractal dimension, the following methods are used for comparative experiments.

\begin{itemize}
	
	\item FSA+RS: It uses Artificial Fish Swarms Algorithm(FSA) combined with Rough Set (RS)\cite{reddy2020hybrid}.
	
	\item AFSA+RS: It uses Improved Artificial Fish Swarm Algorithm(AFSA) combined with RS \cite{2016A}.
	
	\item FSA+RS: It uses Binary Glowworm Swarm Optimization Algorithm(BGSO) \cite{K2009Glowworm} combined with Fractal Dimension (FD).

	\item IDGSO+MFD: It uses an Improved Discrete Glowworm Swarm algorithm (IDGSO) combined with multifractal (MFD). 
	
		\item FWA+FD: It uses the FWA \cite{li2019comprehensive} combined with FD.
		
	\item LFWA+FD (Ours): It uses the LFWA combined with FD.
\end{itemize}

The experimental results in Table 3 show that our "LFWA+FD" method performs best when compared with other comparative methods. Compared with other methods, our "LFWA+FD" method improves the accuracy on Dataset1 by +9.89\%, +6.91\%, +16.27\%, +16.27\%, and 3.29\%, respectively. Similarly, the accuracy of our "LFWA+FD" method on Dataset2 is improved by +7.76\%, +5.91\%,+5.91\%,  +5.81\%, and +2.76\%, respectively. Therefore, these comparative experimental results support the effectiveness of our proposed method.

\subsection{Ablation Study}

To analyze the effectiveness of the LFWA-based feature selection method under the fractal dimension constraint, we did ablation experiments. M1 is our "LFWA+FD" method without LFWA and fractal dimension method. That is, M1 only has a KNN classifier and cannot perform feature dimensionality reduction, so it uses the original features as the input to the classifier. Our "LFWA+FD" method can implement feature dimensionality reduction, so it uses the features after dimensionality reduction as the input of the classifier. The M2 method removes the fractal dimension and retains only LFWA and KNN.The results of the ablation experiments are shown in Table 4. The experimental results show that the performance of our method on Dataset1 is inferior to that of M1 and M2, but our "LFWA+FD" method uses fewer features. On Dataset2, our "LFWA+FD" method outperforms M1 by +4.85\%. But our model performs worse than M2 by -4.04\%. M2 uses features with more dimensions to obtain more information for model inference, so it performs better. In short, our "LFWA+FD" method transmits fewer features for model inference to reduce time cost, which has important application significance in real-time scenarios.

\section{Conclusions}

To address the drawbacks of the original fireworks algorithm with many parameters and increased tuning cost, we propose the LFWA method and apply it to solve the feature selection problem after the robot acquires the data in different scenes. The comparison experimental results of two publicly available datasets from UCI show that the performance of the proposed LFWA method is better than the comparison methods. Moreover, the proposed method is not only effective in searching for the ideal feature subset, but also reduces the data dimensionality and a large amount of noise present in the original data, which in turn ensures the best overall performance of the framework. In our future research, we will focus on the research work of the fireworks algorithm in feature noise reduction and feature fusion of multimodal data collected by different sensors in different scenarios.



%

\bibliographystyle{IEEEtran} 	
\bibliography{reference}

\end{document}